\documentclass[acmtog, authorversion]{acmart}

\usepackage{booktabs}
\usepackage{graphicx}
\usepackage{enumitem}
\usepackage{soul}
\usepackage{dsfont}
\usepackage{gensymb}
\usepackage{microtype}
\usepackage{pifont}
\usepackage{threeparttable}
\usepackage{collect}
\usepackage{xspace}
\usepackage{xfrac}
\usepackage[capitalise]{cleveref}
\usepackage{natbib}
\usepackage{array}
\usepackage{multirow}
\usepackage{wrapfig}
\usepackage{url}

\usepackage{pifont}
\newcommand{\cmark}{\checkmark}%
\newcommand{\xmark}{\scalebox{0.85}{\ding{53}}}%

\creflabelformat{equation}{#2\textup{#1}#3}

\DeclareGraphicsExtensions{.png,.jpg,.pdf,.ai,.psd}
\acmJournal{TOG}
\acmSubmissionID{321}

\usepackage{xr}
\makeatletter
\newcommand*{\addFileDependency}[1]{
  \typeout{(#1)}
  \@addtofilelist{#1}
  \IfFileExists{#1}{}{\typeout{No file #1.}}
}
\makeatother

\newcommand{\refFig}[1]{Fig.~\ref{fig:#1}}

\usepackage[bold=0.55]{xfakebold}
\newcommand{\winner}[1]{\setBold #1\unsetBold}

\usepackage{icomma}

\newcommand{\revision}[1]{\textcolor{black}{#1}}

\DeclareMathOperator\supp{supp}

\newcommand{\eg}{e.g.,\ }
\newcommand{\ie}{i.e.,\ }

\definecollection{mymaths}

\newcommand{\mymath}[2]{
    \newcommand{#1}{\TextOrMath{$#2$\xspace}{#2}}
    \begin{collect}{mymaths}{}{}{}{}
    #1
    \end{collect}
}

\mymath{\field}{f}
\mymath{\kernel}{g}
\mymath{\polykernel}{\hat{g}}
\mymath{\posOneD}{x}
\mymath{\pos}{\mathbf{\posOneD}}
\mymath{\offset}{\boldsymbol\tau}
\mymath{\indim}{{d_{\mathrm{in}}}}
\mymath{\outdim}{{d_{\mathrm{out}}}}
\mymath{\kerneldim}{d}
\mymath{\pdf}{p}
\mymath{\mcsamplecount}{N}
\mymath{\intcount}{n}
\mymath{\diraccount}{m}
\mymath{\postmod}{\rho}
\mymath{\dirac}{\delta}
\mymath{\ramp}{\dirac}
\mymath{\minimalKernel}{h_\intcount}
\mymath{\network}{\hat\field^\intcount}
\mymath{\diracpos}{\mathbf{x}^{(i)}}
\mymath{\diracmag}{w^{(i)}}
\mymath{\transformation}{\mathsf T}
\mymath{\jointcount}{J}

\begin{document}

\setcopyright{acmlicensed}
\acmJournal{TOG}
\acmYear{2023} 
\acmVolume{42} 
\acmNumber{6} 
\acmArticle{206} 
\acmMonth{12} 
\acmPrice{15.00}
\acmDOI{10.1145/3618340}

%----------------------------------------------------------------



%----------------------------------------------------------------
\title{Neural Field Convolutions by Repeated Differentiation}

\author{Ntumba Elie Nsampi}
\affiliation{%
	\institution{MPI Informatik}
	\country{Germany}
}
\email{nnsampi@mpi-inf.mpg.de}

\author{Adarsh Djeacoumar}
\affiliation{%
	\institution{MPI Informatik}
	\country{Germany}
}
\email{adjeacou@mpi-inf.mpg.de}

\author{Hans-Peter Seidel}
\affiliation{%
	\institution{MPI Informatik}
	\country{Germany}
}
\email{hpseidel@mpi-sb.mpg.de}

\author{Tobias Ritschel}
\affiliation{%
	\institution{University College London}
	\country{United Kingdom}
}
\email{t.ritschel@ucl.ac.uk}

\author{Thomas Leimkühler}
\affiliation{%
	\institution{MPI Informatik}
	\country{Germany}
}
\email{thomas.leimkuehler@mpi-inf.mpg.de}

\begin{abstract}

Neural fields are evolving towards a general-purpose continuous representation for visual computing.
Yet, despite their numerous appealing properties, they are hardly amenable to signal processing.
As a remedy, we present a method to perform general continuous convolutions with general continuous signals such as neural fields.
Observing that piecewise polynomial kernels reduce to a sparse set of Dirac deltas after repeated differentiation, we leverage convolution identities and train a repeated integral field to efficiently execute large-scale convolutions.
We demonstrate our approach on a variety of data modalities and spatially-varying kernels.
\end{abstract}

\begin{teaserfigure}
	\includegraphics[width=\textwidth]{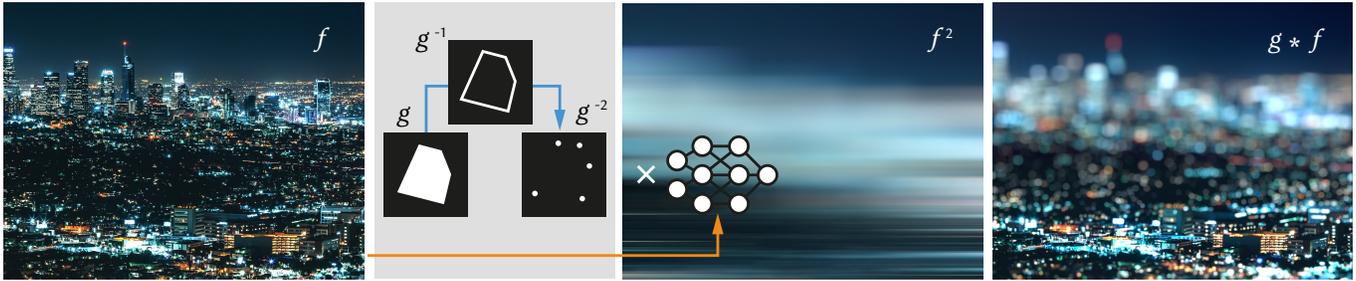}
    %\vspace{-0.7cm}
	\caption{We introduce an algorithm to perform efficient continuous convolution of neural fields $f$ by piecewise polynomial kernels $g$.
    The key idea is to convolve the sparse repeated derivative of the kernel ($g^{-n}$) with the repeated antiderivative of the signal ($f^n$).}
	\label{fig:teaser}
	\Description[TeaserFigure]{TeaserFigure}
\end{teaserfigure}

\maketitle

\section{Introduction}
\label{sec:introduction}

Neural fields have recently emerged as a powerful way of representing signals and have witnessed widespread adoption in particular for visual data \cite{xie2022neural,tewari2022advances}.
Also referred to as implicit or coordinate-based neural representations, neural fields typically use a multi-layer perceptron (MLP) to encode a mapping from coordinates to values.
This representation is universal and allows to capture a multitude of modalities, such as mapping from 2D location to color for images \cite{stanley2007compositional}, from 3D location to the signed distance to a surface for geometry \cite{park2019deepsdf}, from 5D light field coordinates to emitted radiance of an entire scene \cite{mildenhall2020nerf}, and many more.

The appealing properties of neural fields are three-fold:
First, they represent signals in a continuous way, which is a good fit for the mostly continuous visual structure of our world.
Second, they are compact, since they encode complex signals into a relatively small number of MLP weights \cite{dupont2021coin}, while adapting well to local signal complexities.
Third, they are easy to optimize by construction.
Taking all of these properties together, it comes at no surprise that neural fields are rapidly evolving towards a general-purpose data representation \cite{dupont2022data}.
However, to be a true alternative to established specialized representations such as pixel arrays, meshes, point clouds, etc., neural fields are still lacking in a fundamental aspect: 
They are hardly amenable to \emph{signal processing} \cite{xu2022signal,yang2021geometry}.
As a remedy, in this work, we propose a general framework to apply a core signal processing technique to neural fields: convolutions. 

The versatility and expressivity of neural representations have evolved significantly over the last couple of years, mostly due to advances in architectures and training methodologies \cite{sitzmann2020implicit,mueller2022instant,tancik2020fourier,hertz2021sape}.
However, at their core, neural fields only support \emph{point samples}.
This is sufficient for point operations, such as the remapping of input coordinates, \eg for the purpose of deformations \cite{park2021nerfies,tretschk2021non,yuan2022nerf,kopanas2022neural}, or the remapping of output values \cite{vicini2022differentiable}.
In contrast, a convolution requires the \emph{continuous integration} of values over coordinates weighted by a continuous kernel.

Aggregation in neural fields can be approximated using either discretization followed by cubature, or Monte Carlo sampling, resulting in excessive memory requirements and noise, respectively.
Another solution is to consider a narrow parametric family of kernels and train the field using supervision on filtered versions of the signal \cite{barron2021mipnerf}.
Representation and learned convolution operation can be explicitly disentangled using higher-order derivatives \cite{xu2022signal}, but this comes at the cost of only supporting small spatially invariant kernels.
AutoInt~\cite{lindell2021autoint} performs analytic integration using automatic differentiation, but only considers unweighted integrals.
We advance the state of the art by presenting a method to \emph{efficiently} perform \emph{general, large-scale, spatially-varying} convolutions \emph{natively} in neural fields.

In our approach, we consider neural fields to be convolved with piecewise polynomial kernels, which reduce to a sparse set of Dirac deltas after repeated differentiation \cite{heckbert1986filtering}.
Combining this insight with convolution identities on differentiation and integration, our approach requires only a small number of \revision{point} samples from a neural integral field to perform an exact continuous convolution, independent of kernel size.
This integral field needs to be trained in a specific way, supervised via continuous higher-order finite differences, corresponding to a minimal kernel of a certain polynomial degree.
Once trained, our neural fields are ready to be convolved with \emph{any} piecewise polynomial kernel of that degree.

We showcase the generality and versatility of our approach using different modalities, such as images, videos, geometry, character animations, and audio, all natively processed in a neural field representation.
Further, we demonstrate (spatially-varying) convolutions with a variety of kernels, such as smoothing and edge detection of different shapes and sizes.
In summary, our contributions are:
\begin{itemize}
    \item A principled and versatile framework for performing convolutions in neural fields.
    \item Two novel enabling ingredients: An efficient method to train a repeated integral field, and the optimization of continuous kernels that are sparse after repeated differentiation.
    \item The evaluation of our framework on a range of modalities and kernels.
\end{itemize}

\section{Related Work}

\begin{figure}
    \includegraphics[width=0.99\linewidth]{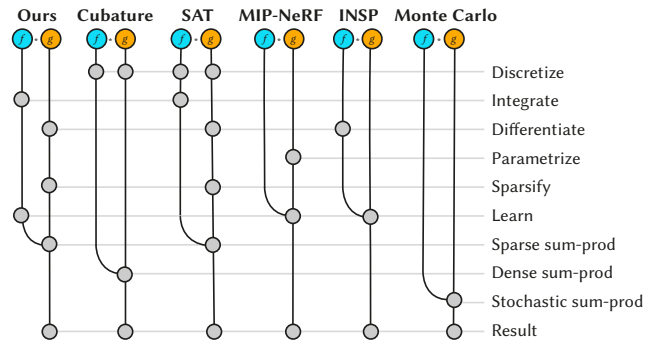}
    %\vspace{-0.25cm}
	\caption{
{The landscape of convolution methods as combinations of different operations applied to a signal and kernel \emph{(top)}, leading to a result \emph{(bottom)}}.
 	}
	\label{fig:alternatives}
\end{figure}

% \begin{table}
%     \centering
%      \caption{Different convolution methods. $n$ is the size of the filter, $m$ the size of the signal (samples or weights to represent it),  and $d$ the signal dimension.}
%      \vspace{-0.3cm}
%     \begin{tabular}{llccc}
%          &
%          \multicolumn1c{Time}&
%          \multicolumn1c{Spat. vary}&
%          \multicolumn1c{Noisy}&
%          \multicolumn1c{Cont.}\\
%          \toprule
%          Classic&
%          $O(m\times n^d)$&
%          \cmark&
%          \xmark&
%          \xmark\\
%          Fourier&
%          $O(m\times \log(m)\times d)$&
%          \xmark&
%          \xmark&
%          \xmark\\
%          Monte Carlo&
%          $O(m \times n)$&
%          \cmark&
%          \cmark&
%          \cmark\\
%          SAT&
%          $O(m\times d)$&
%          \cmark&
%          \xmark&
%          \xmark\\
%          Mip-NeRF&
%          $O(m)$&
%          \cmark&
%          \xmark&
%          \cmark\\
%          INSP&
%          $O(m)$&
%          \xmark&
%          \xmark&
%          \cmark\\
%          \textbf{Ours}&
%          $O(m)$&
%          \cmark&
%          \xmark&
%          \cmark\\
%          \bottomrule
%     \end{tabular}
%     \label{tab:my_label}
% \end{table}

The convolution of a signal, eventually in some higher dimension, with a kernel is a central operation in modern signal processing \cite{zhang2022modern}.
In this work, we consider a slightly generalized form of convolution, where a kernel \emph{varies spatially} across the signal.
Refer to appendix A in the supplemental document for a tabulated overview of the different solutions discussed next.

\paragraph{Discrete}
For discrete signals and kernels (\revision{Cubature and SAT} in \refFig{alternatives}), most convolutions are based on cubature, \revision{\ie a dense sum-product operation} across all dimensions.
This, unfortunately, does not scale to larger kernels or higher dimensions but allows spatially-varying kernels.
A common acceleration is the \emph{Fourier transform} \cite{brigham1988fast}, which also requires time and space to perform and store the transformed signal.
Most of all, it requires the kernel to be spatially invariant.
\revision{As a remedy, certain spatially-varying convolutions can be realized using spatially-varying combinations or transformations of stationary filters \cite{freeman1991design,fournier1988constant,Mitchel:2020:ESA}.}
For a large class of filters, \emph{pyramidal} schemes \cite{lance1983pyramidal,farbman2011convolution} can be a solution, but require additional memory.
The key idea is that intermediate pyramid values store a partial aggregate of the signal.
Techniques that store integrals without reducing the resolution are called \emph{summed-area tables} (SAT) or \emph{integral images} \cite{crow1984summed,viola2001rapid}.
Notably, SATs and their variants allow efficient spatially-varying convolution by considering differentiated kernels \cite{heckbert1986filtering,simard1998boxlets,Leimkuhler2018}.
This efficiency comes from the fact that the differentiated kernel is sparse (``Sparsify'' \revision{for SAT} in \refFig{alternatives}) and the SAT only needs to be evaluated at very few locations.
Our approach will take this idea to the continuous neural domain.

\paragraph{Continuous}
Convolution becomes more challenging, if the signal, the kernel, or both are continuous.
\emph{Monte Carlo} methods, that straightforwardly sample signal and kernel randomly and sum the result (\revision{Monte Carlo} in \refFig{alternatives}) can handle this case.
These scale very well to high dimensions, but at the expense of noise that only vanishes with many samples, even when specialized blue noise \cite{singh2019analysis} or low-discrepancy \cite{niederreiter1992low,sobol1967distribution} samplers are used.
Similar to our approach, the use of derivatives has been shown to be beneficial \cite{kettunen2015gradient}.
Practical unstructured convolution \cite{hermosilla2018monte,wang2018deep,vasconcelos2022cuf,shocher2020discrete}
 does away with cubature and evaluates the product of kernel and signal only at specific sparse positions such as the points in a point cloud.
 Our approach does not rely on random sampling but works directly on a continuous signal.

\paragraph{Neural}
It has recently been proposed to replace discrete representations with continuous neural networks, so-called \emph{neural fields} \cite{tewari2022advances,xie2022neural}.
These have applications in geometry representation \cite{park2019deepsdf}, novel-view synthesis \cite{sitzmann2019srns, mildenhall2020nerf}, dynamic scene reconstruction \cite{park2021nerfies,tretschk2021non,yuan2022nerf}, etc.
Replacing a discrete grid with complex continuous functions requires developing the same operations available to grids \cite{dupont2022data}, including convolutions, as we set out to do in this work.
Early work has been conducted to explore the manifold of all natural neural fields \cite{du2021gem} and to build a generative model of neural fields \cite{dupont2021generative,erkoc2023hyperdiffusion}.
Specialized network architectures allow the decomposition of signals into a discrete set of frequency bands \cite{fathony2020multiplicative,lindell2022bacon,yang2022polynomial}.
Further, limited forms of geometry processing have been considered in this representation \cite{yang2021geometry}.
However, none of them is looking into general, efficient, large-scale, and/or spatially-varying convolutions.

A very specific form of convolution occurs as anti-aliasing or depth-of-field in image-based rendering.
To account for these effects, neural fields can be learned that are conditioned on the parameters of the convolution kernel, such as its bandwidth \cite{barron2021mipnerf,wang2022nerfocus,isaac2022exact,barron2022mipnerf360}.
The network can then be evaluated to directly produce the filtered result, \ie signal representation and convolution operation are intertwined.
This Mip-NeRF-style convolution is in principle applicable to other filters, as long as they can be parametrized to become conditions to input into the network (MIP-NeRF in \refFig{alternatives}).
Unfortunately, this limits the kernels that can be applied to a parametric family that needs to be known in advance.
Further, it significantly increases training time, since kernel parameters act as additional input dimensions to the network.
We use inductive knowledge of integration and differentiation to arrive at a more efficient formulation that generalizes across kernels.

The key to efficient convolutions is a combination of sparsity, differentiation, and integration.
Fortunately, tools to perform integration and differentiation on neural fields are available.
AutoInt~\cite{lindell2021autoint} proposes to learn a neural network that, when automatically differentiated, fits a signal.
By evaluating the original network without differentiation, the antiderivative can be evaluated conveniently.
Unfortunately, this approach does not scale well to the higher-order antiderivatives needed for efficient convolutions, as the size of the derivative graphs grows quickly.
In contrast, our approach leverages higher-order finite differences to train a repeated integral field, which scales with the number of integrations required.

Recently, \citet{xu2022signal} have proposed a method with the same aim as ours (\revision{INSP} in \refFig{alternatives}).
Given a trained neural field, they learn to combine higher-order derivatives of the field to approximate a convolution.
Similar to a Taylor expansion, this requires high-order derivatives to reason about larger neighborhoods and, unfortunately, hence only allows for very small, spatially invariant kernels.

Finally, aggregation in neural fields in the form of range queries has been studied by \citet{spelunking22}.
Their approach allows to retrieve \revision{a conservative estimate of} the field's extrema within a query volume, which unfortunately does not provide enough information to perform accurate continuous convolutions.

\section{Background}
We consider the convolution of arbitrary continuous signals 
$\field \in \mathds{R}^{\indim} \rightarrow \mathds{R}^{\outdim}$  with arbitrary continuous kernels $\kernel \in \mathds{R}^{\kerneldim} \rightarrow \mathds{R}$.
Both inputs and outputs of \field are low- to medium-dimensional.
The signal can be any continuous function, including but not limited to a  neural network.
We assume the kernel has compact but potentially large support.
The kernel does not necessarily extend across \emph{all} input dimensions of \field, \ie $\kerneldim \leq \indim$.
To simplify our exposition, without loss of generality, we assume that the first \kerneldim dimensions of \field correspond to the filter dimensions of \kernel.
Further, we allow \kernel to vary for different locations in the input space. An example of this setup is a space-time signal \field encoding an RGB ($\outdim=3$) video with two spatial and a temporal dimension ($\indim=3$), to be convolved with a kernel \kernel that applies a foveated blur to each time slice of the video ($\kerneldim = 2$).
In the following derivations, we assume a spatially invariant kernel for ease of notation.
Sec.~\ref{sec:efficient-neural-field-convolutions} explains how our method can be easily extended to the spatially-varying case.

Formally, we seek to carry out the continuous convolutions
\begin{equation}
\label{eq:basic_conv}
    (\field * \kernel) (\pos)
    =
    \int_{\mathds{R}^\kerneldim} 
    \field(\pos - \offset) \kernel(\offset) 
    \mathrm{d} \offset.
\end{equation}
This integral operation does not have a closed-form solution for all but the most constrained sets of signals and/or kernels.
In particular, it is unclear how this continuous operation can be applied to generic neural fields, which naturally only support point samples.
The typical solution for these integrals is numerical approximation:
For low-dimensional integration domains, quadrature rules are feasible, while the scalable gold standard in higher dimensions is Monte Carlo integration.
The latter proceeds by sampling the integration domain and approximating the integral by a weighted sum of integrand evaluations:
\begin{equation}
\label{eq:mc_conv}
    (\field * \kernel) (\pos)
    =
    \mathbb{E}_{\offset}
    \left[
    \field(\pos - \offset) \kernel(\offset)
    \right]
    \approx
    \frac{1}{\mcsamplecount}
    \sum_{\offset \sim \pdf}
    \frac{
        \field(\pos - \offset) \kernel(\offset)
    }{
        \pdf(\offset)
    },
\end{equation}
where $\mathbb{E}$ is the expectation and $\offset$ are now random samples drawn from the probability density function \pdf. Unfortunately, a high number \mcsamplecount of samples is required for large kernels \kernel and/or high-frequency signals \field, rendering this approach inefficient.

In the following, we develop a method that performs continuous convolutions in the form of Eq.~\ref{eq:basic_conv}, while only requiring a very low number of network evaluations, independent of the kernel size.

\begin{figure*}[h!]
	\includegraphics[width=0.99\linewidth]{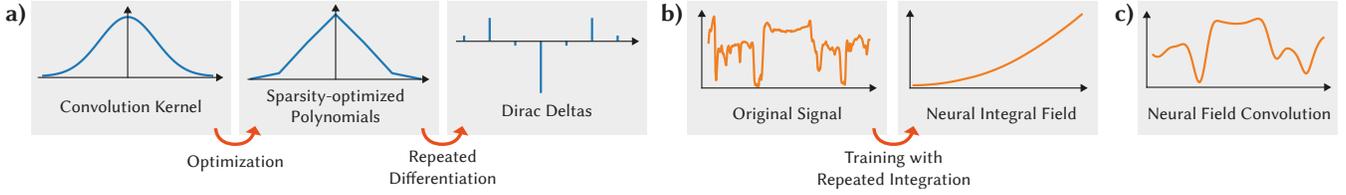}
    %\vspace{-0.4cm}
	\caption{
	Overview of our approach.
    \emph{a}) Given an arbitrary convolution kernel, we optimize for its piecewise polynomial approximation, which under repeated differentiation yields a sparse set of Dirac deltas.
    \emph{b}) Given an original signal, we train a neural field that captures the repeated integral of the signal.
    \emph{c}) The continuous convolution of the original signal and the convolution kernel is obtained by a discrete convolution of the sparse Dirac deltas from \emph{a}) and corresponding sparse samples of the neural integral field from \emph{b}).
	}
	\label{fig:overview}
\end{figure*}

\section{Method}

We efficiently convolve a continuous signal \field with a continuous kernel \kernel by approximating \kernel with a piecewise polynomial function, which becomes sparse after repeated differentiation. 
Our approach requires the evaluation of the repeated integral of \field at a sparse set of sample positions dictated by the differentiated kernel (Sec.~\ref{sec:method_convolution}), leading to a substantial speed-up of the convolution operation. 
This general approach has first been studied by \citet{heckbert1986filtering} in the context of discrete representations.
Our method lifts the idea to the continuous neural setting by optimizing for a sparse differential representation of \kernel (Sec.~\ref{sec:method_kernel}), and obtaining the repeated continuous integral of \field by supervising on a minimal kernel using higher-order finite differences (Sec.~\ref{sec:method_integral_field}).
Once trained, any sparsity-optimized convolution kernel can be applied to \field efficiently, without requiring additional input parameters.
Our method supports spatially-varying convolutions in the form of continuously transformed kernels, leveraging the continuous nature of the representation. Fig.~\ref{fig:overview} gives an overview of our approach.

%==============================================================

\subsection{Convolution by Repeated Differentiation}
\label{sec:method_convolution}

Our approach requires two conceptual ingredients:
First, the convolution operation in Eq.~\ref{eq:basic_conv} reduces to a discrete sum if \kernel consists of only Dirac deltas (Sec.~\ref{sec:dirac-kernels}).
Second, the right-hand side of Eq.~\ref{eq:basic_conv} can be transformed using repeated differentiation and integration (Sec.~\ref{sec:convolutions-with-diff-int}).
Putting both ingredients together leads to an efficient discrete formulation of the continuous convolution operation (Sec.~\ref{sec:efficient-neural-field-convolutions}) involving a repeated integral field and a sparse differential kernel consisting of Dirac deltas \cite{heckbert1986filtering}.

\subsubsection{Dirac Kernels}
\label{sec:dirac-kernels}

As our first ingredient, consider a kernel \kernel that is non-zero only at a small set of \diraccount locations in $\mathds{R}^\kerneldim$, \ie \kernel is
\begin{equation}
\label{eq:dirac_kernel}
    \kernel(\pos)
    =
    \sum_{i=1}^\diraccount 
    \delta ( \pos - \diracpos )
    \diracmag,
\end{equation}
a sum of Dirac deltas $\delta$, where $\diracpos \in \mathds{R}^d$ denotes the location and $\diracmag \in \mathds{R}$ the magnitude\footnote{Technically, $\delta(\mathbf{0})=\infty$. But since a Dirac delta integrates to one, in the context of continuous convolutions, we refer to \diracmag as ``magnitudes'' nevertheless.}
of the $i$'th impulse.
Then, by the \emph{sifting} property of Dirac deltas, Eq.~\ref{eq:basic_conv} simplifies to
\begin{equation}
\label{eq:dirac_conv}
    (\field * \kernel) (\pos)
    =
    \sum_{i=1}^\diraccount
    \field (\pos - \diracpos) \diracmag,
\end{equation}
\ie we have reduced the computation from a continuous integral to a discrete sum -- an efficient operation if \diraccount is small.

\subsubsection{Convolutions with Differentiation and Integration}
\label{sec:convolutions-with-diff-int}

Our second ingredient is the following identity:
\begin{equation*}
    \field * \kernel 
    = 
    \left(
    \int \!\! \field \mathrm{d}\pos_i
    \right)
    *
    \left(
    \frac{\partial}{\partial \pos_i} \kernel
    \right),
\end{equation*}
\ie in order to convolve \field with \kernel we might as well convolve the antiderivative of \field with the corresponding derivative of \kernel.
Applying this principle repeatedly yields \revision{\cite{heckbert1986filtering, perlin_personal}}
\begin{equation}
\label{eq:diff_conv}
    \field * \kernel 
    = 
    \underbrace{
    \left(
    \int^\intcount \!\!\!\! \ldots \! \int^\intcount
    \!\! \field 
    \mathrm{d}\pos_1^{\intcount} \ldots \mathrm{d}\pos_{\kerneldim}^{\intcount}
    \right)
    }_{\field^\intcount}
    *
    \underbrace{
    \left(
    \frac
        {\partial^{\kerneldim \intcount}}
        {\partial \pos_1^{\intcount} \ldots \partial \pos_{\kerneldim}^{\intcount}}
        \kernel
    \right)
    }_{\kernel^{-\intcount}}
    .
\end{equation}
Here, we sequentially differentiate \kernel \intcount times with respect to \emph{each} of its dimensions. 
We denote this multidimensional repeated derivative as
$\kernel^{-\intcount}$.
For equality in Eq.~\ref{eq:diff_conv} to hold, this pattern is mirrored for \field, replacing differentiations with antiderivatives, where superscripts $\intcount$ denote repeated integrations along the individual dimensions.
We denote the repeated multidimensional antiderivative of \field as 
$\field^{\intcount}$.
We refer to \citet{heckbert1986filtering} for a proof of Eq.~\ref{eq:diff_conv}.
\revision{Notice that input and output dimensions of \field and \kernel do not change after integration and differentiation.}

\subsubsection{Efficient Neural Field Convolutions}
\label{sec:efficient-neural-field-convolutions}

The central idea of our approach is to combine both ingredients presented above for the case of piecewise polynomial kernels \polykernel.
Concretely, we observe that piecewise polynomial functions turn into a sparse set of Dirac deltas after repeated differentiation \cite{heckbert1986filtering} (Fig.~\ref{fig:kernel_differentiation}), \ie
$\polykernel^{-\intcount}$
reduces to the form of Eq.~\ref{eq:dirac_kernel}.
This implies that Eq.~\ref{eq:dirac_conv} can be used to perform a convolution with this kernel.
Combining Eq.~\ref{eq:dirac_conv} and Eq.~\ref{eq:diff_conv}, our final convolution operation reads
\begin{equation}
\label{eq:our_conv}
\boxed{
    (\field * \polykernel) (\pos)
    =
    \sum_{i=1}^\diraccount
    \field^{\intcount}
    (\pos - \diracpos) 
    \diracmag.
    }
\end{equation}
Notice that this formulation requires only \diraccount evaluations of the repeated integral of \field at locations dictated by the Dirac deltas of the differentiated kernel to yield the same result as the equivalent continuous convolution in Eq.~\ref{eq:basic_conv}.

The number of integrations and differentiations \intcount directly depends on the desired order of the kernel polynomials, as detailed in Sec.~\ref{sec:method_kernel}.
A disk-shaped kernel simulating thin-lens depth of field in an image can be approximated well using a piecewise \emph{constant} function (corresponds to $\intcount=1$), while a Gaussian might require a piecewise \emph{quadratic} approximation (corresponds to $\intcount=3$) to yield high-quality results with a low number of Dirac deltas.

Notice that our approach allows us to realize \emph{spatially-varying} convolutions as well:
The evaluation of Eq.~\ref{eq:our_conv} is independent for different evaluation locations \pos.
Therefore, we can make the choice of the convolution kernel \polykernel a function of \pos itself.
In Sec.~\ref{sec:kernel_transformations} we give details on how to obtain continuous parametric kernel families.

In summary, our method requires two components:
(\emph{i}) A piecewise polynomial kernel that results in a sparse set of Dirac deltas after repeated differentiation, and
(\emph{ii}) an efficient way to obtain and evaluate the repeated multidimensional integral of a continuous signal.
These are detailed in Sec.~\ref{sec:method_kernel} and Sec.~\ref{sec:method_integral_field}, respectively.

%==============================================================

\subsection{Sparse Differential Kernels}
\label{sec:method_kernel}

\begin{figure}
	\includegraphics[width=0.99\linewidth]{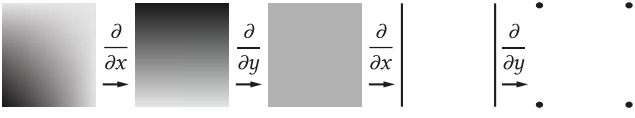}
    %\vspace{-0.3cm}
	\caption{
	Repeated differentiation of a bilinear patch ($\kerneldim=\intcount=2$). After the first differentiation, the patch exhibits linear variation only along the vertical dimension. After subsequent differentiations, we obtain a constant patch, two vertical lines, and, finally, four Dirac deltas.
	}
	\label{fig:kernel_differentiation}
\end{figure}

\begin{figure}[h!]
	\includegraphics[width=0.99\linewidth]{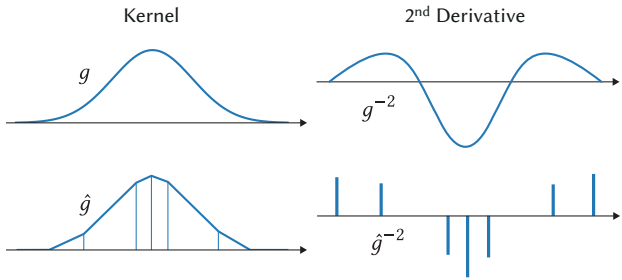}
    %\vspace{-0.4cm}
	\caption{
	  Kernel representation in 1D for the case $\intcount=2$, \ie a piecewise linear function.
    \emph{Top row}: The original continuous kernel \kernel has a continuous second derivative.
    \emph{Bottom row}: We approximate \kernel with a piecewise linear function $\hat{\kernel}$, which reduces to a sparse set of Dirac deltas in its second derivative.
	}
	\label{fig:kernel_opt}
\end{figure}

Our approach requires a kernel \kernel which, after repeated differentiation, results in a sparse set of Dirac deltas with positions \diracpos and magnitudes \diracmag:
\begin{equation}
\label{eq:diff_kernel}
    \kernel^{-\intcount}
    (\pos)
    =
    \sum_{i=1}^\diraccount 
    \delta ( \pos - \diracpos )
    \diracmag.
\end{equation}
This property is satisfied for piecewise polynomial kernels of degree 
$\intcount-1$, which reduce to Dirac deltas positioned at the junctions between the segments after \intcount differentiations per dimension \cite{heckbert1986filtering} (Fig.~\ref{fig:kernel_differentiation}).
Thus, given a kernel \kernel, we seek to find its optimal piecewise polynomial approximation $\hat{\kernel}$ adhering to a user-specified budget of \diraccount Dirac deltas (Fig.~\ref{fig:kernel_opt}).

\begin{figure}[h!]
	\includegraphics[width=0.99\linewidth]{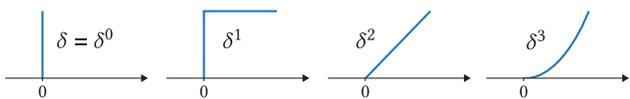}
    % \vspace{-0.3cm}
	\caption{
	  1D ramps of different orders (redrawn from \citet{heckbert1986filtering}). Each ramp is the antiderivative of its predecessor.
	}
	\label{fig:ramps}
\end{figure}

To parameterize $\hat{\kernel}$, we utilize the linear structure of Eq.~\ref{eq:diff_kernel} and the linearity of differentiation: 
We consider the \kerneldim-dimensional \intcount-fold repeated antiderivative of the Dirac delta function
\begin{equation*}
    \ramp^\intcount(\pos) = 
    \begin{cases}
        \frac{
            \prod_{i=1}^{\kerneldim} \pos_i^{\intcount-1}
        }{
            (\intcount-1)^\kerneldim !
        } 
        & \min_i \pos_i \geq 0 \\
        0 & \text{else}
    \end{cases}
\end{equation*}
which is referred to as the \intcount'th-order \emph{ramp} (Fig.~\ref{fig:ramps}).
We now write our polynomial kernel $\hat{\kernel}$ as a linear combination of shifted ramps:
\begin{equation}
\label{eq:kernel_superposition}
    \hat{\kernel}(\pos)
    =
    \sum_{i=1}^\diraccount 
    \ramp^\intcount ( \pos - \diracpos )
    \diracmag.
\end{equation}
Please note that \cref{eq:diff_kernel} is the \intcount'th derivative of \cref{eq:kernel_superposition} by construction.
Thus, we have established a parameterization of a $C^{\intcount-2}$-continuous piecewise polynomial kernel $\hat{\kernel}$, from which we can directly read off Dirac delta positions and magnitudes.

We now optimize the following objective:
\begin{equation}
    \min_{\diracpos, \diracmag}
    \left[
        \mathbb{E}_{\pos \in \mathds{R}^\kerneldim}
        \left[
        \left\lVert
            \kernel(\pos) - \hat{\kernel}(\pos)
        \right\lVert_2^2
        \right]
        +
        \lambda
        \left|
            \sum_{i=1}^\diraccount
            \diracmag
        \right|
    \right].
\end{equation}
The first term encourages the solution to be close to the reference kernel on the entire domain. 
The second term steers the optimization to prefer solutions where the Dirac magnitudes sum to zero, which effectively enforces the kernel to be compact.

\subsubsection{Optimization}
\label{sec:kernel_optimization}
We initialize the ramp positions \diracpos on a regular grid and their magnitudes \diracmag to zero. 
We use the Adam \cite{kingma2014adam} optimizer with standard parameters and set $\lambda=0.1$ in all our experiments.
In each iteration, we uniformly sample the continuous kernel domain.
We observe that when the provided ramp budget \diraccount is too high, the magnitudes of individual ramps will approach zero, further increasing sparsity. 
We capitalize on this fact by monitoring ramp magnitudes in regular intervals.
If an absolute magnitude falls below a small threshold, we remove the ramp from the mixture and continue optimizing.
Separable kernels can be obtained by optimizing the respective 1D filters and combining them with an outer product.
We provide timings of the optimization for different kernels in Supplemental appendix C.

\revision{We note the strong connection to spline-based approximations of functions \cite{ahlberg2016theory}. 
In the low-order regime under the continuity requirements we operate on, we find that our practical stochastic gradient descent-based approach yields high-quality results without the need for more elaborate techniques.}

\subsubsection{Kernel Transformations}
\label{sec:kernel_transformations}
Many applications of convolutions require kernels of different sizes and shapes, in particular in the case of spatially-varying convolutions.
For example, foveated imagery or the simulation of depth-of-field require continuous and fine-grained control over the size of a blur kernel.
Our approach supports on-the-fly kernel transformations without the need to sample and optimize entire parametric families of kernels by leveraging the continuous nature of the kernel and the signal.

Concretely, to continuously shift and (anisotropically) scale an optimized kernel \polykernel using a matrix \transformation, we simply apply \transformation to the Dirac delta positions, \ie
$
\diracpos_\transformation=
\transformation \diracpos
$.
The updated Dirac delta magnitudes are given by
$
\diracmag_\transformation
=
\frac{\diracmag}{\det(\transformation)^{\intcount}}
$.
Thus, we need to run the optimization for a kernel type only once in a canonical position and size, and obtain continuously transformed kernel instances at virtually no computational cost.
Notice that, as a useful consequence, our approach enables continuous scale-space analysis \cite{witkin1987scale,lindeberg2013scale}, as illustrated in supplemental.

%==============================================================

\subsection{Neural Repeated Integral Field}
\label{sec:method_integral_field}

% \begin{figure}
% 	\includegraphics[width=0.99\linewidth]{figures/estimators}
%     \vspace{-0.4cm}
% 	\caption{
% 	 Antiderivative loss estimation:
%     The first row shows an HDR image with four stars.
%     These are four box-peaks in the second row, showing a scanline of that image.
%     To compute any value for the antiderivative in the third row, (dotted line) we need to sum all integrand (signal) values to the left, the entire red area.
%     This will miss peaks.
%     The fourth row shows our alternative loss, that supervises on a convolution with a minimal kernel.
%     It only requires samples in the small, blue area, which will not miss a star.
% 	}
% 	\label{fig:estimators}
% \end{figure}

\begin{figure}
	\includegraphics[width=0.99\linewidth]{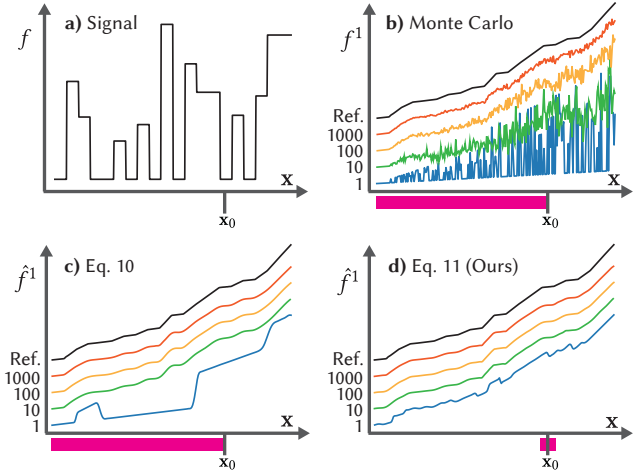}
    %\vspace{-0.2cm}
	\caption{
	 \revision{
  (\emph{a}): Given a signal \field, we are interested in finding its antiderivative $\field^1$. 
  (\emph{b}): A scalable way to obtain the antiderivative is Monte Carlo estimation $\mathbb E_{\offset\geq0}\left[\field(\pos - \offset)\right]$. Increasing the number of samples (different colors) gets us closer to the true solution (black).
  (\emph{c}): Using the estimates from \emph{b}) to supervise the training of a network $\hat\field^1$ as per Eq.~\ref{eq:loss_mc} requires many samples, while the regressed antiderivative remains blurry.
  (\emph{d}): In contrast, our approach only requires a convolution with a small kernel (Eq.~\ref{eq:loss}) to yield high-quality results, including sharp features, with only a low number of Monte Carlo samples. For each method, a pink bar marks the region that needs to be considered for estimating/training the antiderivative value at location $\pos_0$. The different graphs in \emph{b})-\emph{d}) use different constants of integration for improved visualization.
  }}
	\label{fig:sample_count}
\end{figure}

To compute \cref{eq:our_conv}, we need to evaluate $\field^{\intcount}$, the \intcount-th antiderivative of \field as the second ingredient.
We choose to implement $\field^{\intcount}$ as a neural field \network.
Ideally, it would hold that $\network=\field^{\intcount}$.
This might be difficult to achieve without knowing an analytic form of the antiderivative.
We could try Monte Carlo-estimating the antiderivative from the signal, leading to a loss like
\begin{align}
\label{eq:loss_mc}
\mathbb E_{\pos \in \mathbb{R}^{\indim}}
\left[
\left\lVert
\network(\pos)
-
\mathbb E_{\offset\geq0}
\left[
\field(\pos - \offset)
\right]
\right\lVert
\right].
\end{align}
The inner expectation would sum over the entire half-domain $\offset\geq0$, leading to a high variance and a low-quality \network.
\revision{
An example of this is shown in \cref{fig:sample_count}, where the input is a 1D HDR signal (\cref{fig:sample_count}a).
To estimate the antiderivative at $\pos_0$, we have to sample the entire pink area in \cref{fig:sample_count}b, resulting in significant variance, even if the sample count increases.
When using these estimates to train \network, this variance leads to a low-quality, blurry regression (\cref{fig:sample_count}c), as no loss function is known that properly captures the statistical properties of Monte Carlo noise \cite{lehtinen2018noise2noise}.
}

For our purpose, convolution, what really needs to hold, is that 
$
\network*\minimalKernel^{-n}
=
\field*\minimalKernel
$, for any piecewise polynomial kernel \minimalKernel of degree $\intcount-1$.
%This includes also very compact \minimalKernel, so as to prevent the antiderivative network to ``cheat'' by not learning the antiderivative of the signal but the antiderivative of a convolution of the signal.
The resulting loss to achieve this is

\begin{align}
    \label{eq:loss}
    &\mathbb E_{\pos \in \mathbb{R}^{\indim}}
    \left[
    \left\lVert
     \sum_{i=1}^\diraccount 
    \network ( \pos - \diracpos )
    \diracmag
    -
    \mathbb E_
    {
        \revision{\offset \in \supp(\minimalKernel)}
        %\pos^{(1)}\leq
        %\offset<
        %\pos^{(\diraccount)}
    }
    \left[
    \field(\pos-\offset)
    \minimalKernel(\offset)
    \right]
    \right\lVert
    \right]
    ,
\end{align}
where \diracpos and \diracmag are the Dirac delta positions and magnitudes of
$\minimalKernel^{-\intcount}$.
The first term is due to \cref{eq:our_conv} and the second term is a Monte Carlo estimate of the convolved signal.

\begin{figure}
	\includegraphics[width=0.99\linewidth]{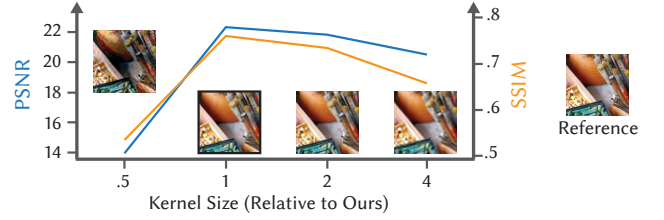}
    %\vspace{-0.3cm}
	\caption{
	 \revision{Antiderivative quality as a function of kernel size used in Eq.~\ref{eq:loss}.
  We measure and display quality by repeated automatic differentiation of the learned antiderivatives. We see that our \emph{minimal kernel} is optimal in terms of quality. Smaller kernels lead to instabilities and larger ones to blur.}
  }
	\label{fig:minimality}
\end{figure}

\revision{
For efficient training and a high-quality antiderivative, it is crucial for \minimalKernel to be very compact:
First, the right part of \cref{eq:loss} becomes a tame interval, significantly reducing variance and thus enabling training with a low number of Monte Carlo samples (Fig.~\ref{fig:sample_count}d).
Second, it prevents \network from ``cheating'' by not learning the antiderivative of the signal but the antiderivative of a convolved signal.
However, we cannot reduce the support of \minimalKernel arbitrarily, as the left part of \cref{eq:loss} tends to result in instabilities, as the distances between the \diracpos shrink.
Fig.~\ref{fig:minimality} illustrates the inherent trade-off of this situation:
There exists a sweet spot for the kernel size, producing the highest-quality antiderivatives.
Smaller kernels lead to training instabilities, larger kernels to blur.
We refer to the optimal solution as the \emph{minimal kernel}.
}

\begin{figure}[h!]
	\includegraphics[width=0.99\linewidth]{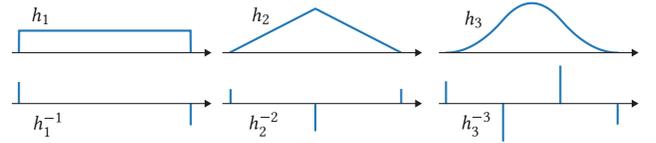}
    % \vspace{-0.3cm}
	\caption{
	  Minimal piecewise polynomial kernels of different degrees (\emph{top row}) and their corresponding Dirac deltas (\emph{bottom row}).
	}
	\label{fig:minimal_kernels}
\end{figure}

\revision{
While we could use any filter shape as minimal kernel \minimalKernel, we choose the $n$-fold convolution of a box with itself (\cref{fig:minimal_kernels}).
This has the advantage that the left part of the loss \cref{eq:loss} becomes a sum over only 
$\diraccount = (\intcount + 1)^\kerneldim$
elements that is efficient to compute, corresponding to higher-order finite differences.
}

%==============================================================

\subsection{Implementation Details}
\label{sec:implementation}

All source code is accessible via \textcolor{blue}{\url{https://neural-fields-conv.mpi-inf.mpg.de}}.
We have implemented our prototype within the PyTorch \cite{paszke2017automatic} environment.
Our integral fields are realized using MLPs, where exact architectures vary slightly depending on the modality to be represented, as detailed in appendix B of the supplemental document.

For training our repeated integral field, we again use the Adam \cite{kingma2014adam} optimizer with standard parameters.
\revision{We handle boundaries by mirror-padding the training signal.}
Considering a unit domain, we train with a size of 0.025 for the minimal kernel \minimalKernel until convergence, followed by a fine-tuning on a kernel of size 0.0125.
\revision{Empirically, we found this size to produce the highest-quality antiderivatives (Fig.~\ref{fig:minimality}).
As the kernel size cannot be further decreased, our repeated integral field \network is a slightly low-pass filtered version of $\field^{\intcount}$, resulting in a lower limit of filter sizes our convolutions can faithfully compute.}
Fortunately, these small filters are highly amenable to efficient Monte Carlo estimation.
Thus, at test time, whenever a kernel size falls below the threshold, we Monte-Carlo-estimate the convolution. % using the same number of samples as we have Diracs for the larger kernels.

\revision{Convolution per Eq.~\ref{eq:our_conv} can be efficiently implemented as an augmented neural field based on \network, by prepending positional offsets \diracpos and appending a linear layer containing \diracmag.} 

%==============================================================

\section{Applications}
\label{sec:applications}

To demonstrate the generality and efficiency of our approach, we consider five signal modalities: images, videos, geometry, character animations, and audio.
An overview of settings for all modalities is given in Tab.~\ref{tab:Applications}.
References are computed using Monte Carlo estimation as per Eq.~\ref{eq:mc_conv} until convergence. 
\revision{Further, we consider the following baselines:}

\begin{table}[]
    \centering
    \caption{Settings for all applications.}
    %\vspace{-0.3cm}
    \renewcommand{\tabcolsep}{0.16cm}    \label{tab:Applications}
    \begin{threeparttable}
  \begin{tabular}{llrlllllr}
&
\multicolumn{3}{c}{Input}&
\multicolumn{4}{c}{Kernel}
\\
\cmidrule(lr){2-4}
\cmidrule(lr){5-9}
\footnotesize{Modality} & \footnotesize{\indim} & \footnotesize{\outdim} & \footnotesize{Format} & \footnotesize{Shape} & \footnotesize{SV\tnote{a}} & \footnotesize{\kerneldim} & \footnotesize{Order} & \footnotesize{\diraccount\tnote{b}} \\
\toprule
Images & 2 & 3 & Grid & Gauss & \xmark & 2 & Linear & 169 \\
& 2 & 3 & Grid & DoG & \xmark & 1 & Linear & 13 \\
& 2 & 3 & Grid & Circle & \cmark & 2 & Const. & 141 \\
Bilat. Images & 3 & 3 & Grid & Gauss & \cmark & 3 & Const. & 343 \\
Video & 3 & 3 & Grid & Tent & \xmark & 1 & Linear & 3 \\
Geometry & 3 & 1 & SDF & Box & \cmark & 3 & Const. & 8 \\
%Rad. Field & 5 & 4 & Field & \unsure{Gauss} & \xmark & 2 & \unsure{Const.} & \unsure{8} \\
Animation & 1 & 69 & Paths & Gauss & \xmark & 1 & Linear & 13 \\
Audio & 1 & 1 & Wave & Box & \xmark & 1 & Const. & 2 \\
\bottomrule
\end{tabular}
\begin{tablenotes}\footnotesize
\item[a] \revision{Experiments with spatially-varying kernels.}
\item[b] \revision{The number of Diracs automatically adapts to the kernel as described in Sec.~\ref{sec:kernel_optimization}.}
\end{tablenotes}
\end{threeparttable}
\end{table}

\paragraph{INSP \cite{xu2022signal}}
\revision{Similar to our approach, INSP relies on higher-order derivatives but uses them in a point-wise fashion to reason about local neighborhoods, reminiscent of a Taylor expansion.
This method has to be trained for each convolution kernel separately, while our integral fields can be used with any kernel of a fixed polynomial degree.
We follow their original implementation and provide all second-order partial derivatives. 
We found that providing more derivatives did not markedly improve their results.}

\paragraph{BACON \cite{lindell2022bacon} and PNF \cite{yang2022polynomial}}
\revision{While our method supports arbitrary filters at test time, both BACON and PNF are limited to a discrete cascade of $k$ specific and fixed intermediate network outputs with different frequency contents, \ie a set of pre-filtered versions of the signal.
To compare ours to BACON and PNF, we approximate the convolution with an arbitrary filter as a linear combination of their intermediate outputs:
Given the reference result, we optimize for a set of weights that when multiplied with corresponding intermediate outputs is closest to the target.
Notice that this procedure requires a reference, while ours does not.
For PNF, we observed that finer-scale intermediate outputs are not zero-mean. 
We compensate for this by subtracting the mean from all intermediate outputs and adding the sum of these means to the coarsest output.
This way, finer levels only add higher-frequency details to the solution, but no global color shifts.
}

%==========================================

\begin{figure}[h!]
	\includegraphics[width=0.99\linewidth]{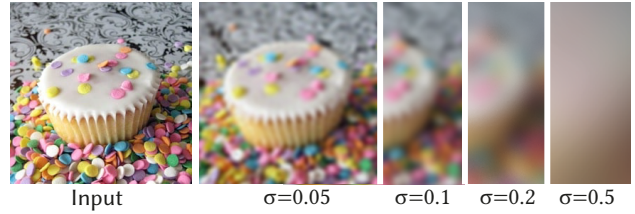}
        %\vspace{-.3cm}
	\caption{
	 Gaussian 2D image blur of the input signal, with increasing bandwidth.
        It can be seen how our approach works also for large kernels.
	}
	\label{fig:gauss_result}
\end{figure}

\begin{figure}[h!]
	\includegraphics[width=0.99\linewidth]{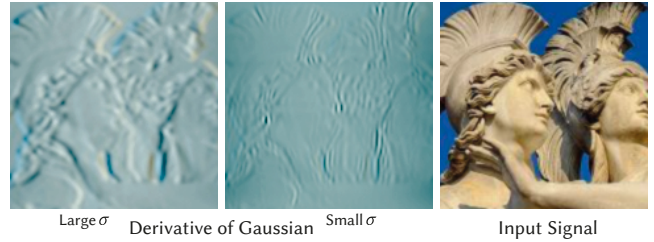}
        %\vspace{-.3cm}
	\caption{
	   Derivative-of-Gaussian filtering 2D result. Note, that our approach supports such non-convex filters, producing signed results.
	}
	\label{fig:dog_result}
\end{figure}

\begin{figure*}[h!]
	\includegraphics[width=0.99\linewidth]{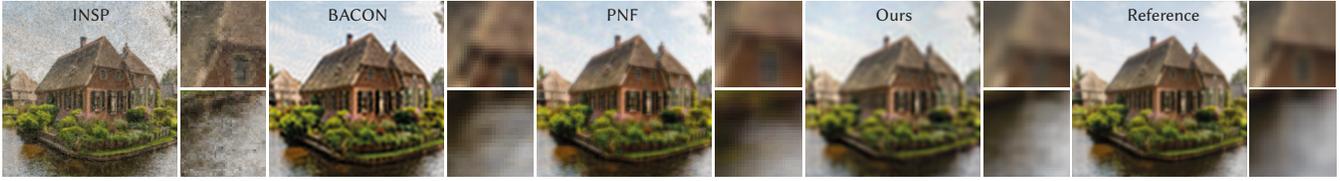}
        %\vspace{-.3cm}
	\caption{
	 A qualitative comparison between INSP \cite{xu2022signal}, BACON \cite{lindell2022bacon},
  PNF \cite{yang2022polynomial}, and our approach using a Gaussian kernel.
        INSP approach does hardly perform any filtering, when the kernel is larger, such as our approach supports, but suffers from noise.
        BACON fairs better, but shows ringing at high-contrast edges such as the house against the sky, while PNF also struggles to produce low-pass filtering results.
	}
	\label{fig:insp_comparison}
\end{figure*}

\subsection{Images}
\label{sec:images}

In this application, we consider (high dynamic range) RGB images $\field \in \mathds{R}^2 \rightarrow \mathds{R}^3$ as signals.

\paragraph{Linear filtering}
We show results for a Gaussian image filter in \cref{fig:gauss_result} and a derivative-of-Gaussian filter in \cref{fig:dog_result}.
More results can be found in supplemental.
\cref{tab:comparison} and \cref{fig:insp_comparison} evaluate image quality on a set of 50 images.
Since no other method can produce competitive results for large kernels, to facilitate an insightful analysis, we limit our numerical evaluation to the small-kernel regime.
We see that for very small kernels, BACON and PNF tend to produce higher-quality results, while we significantly outperform all methods as the kernel size increases.

% \begin{figure}[h!]
% 	\includegraphics[width=0.99\linewidth]{figures/GaussResult}
%         \vspace{-.3cm}
% 	\caption{
% 	 Gaussian 2D image blur of the input signal, with increasing bandwidth.
%         It can be seen how our approach works also for large kernels.
%         References look visually indistinguishable, as seen from the supplemental.
% 	}
% 	\label{fig:gauss_result}
% \end{figure}

% \begin{figure}[h!]
% 	\includegraphics[width=0.99\linewidth]{figures/DoGResult.ai}
%         \vspace{-.3cm}
% 	\caption{
% 	   Derivative-of-Gaussian filtering 2D result. Note, that our approach supports such non-convex filters, producing signed results.
% 	}
% 	\label{fig:dog_result}
% \end{figure}

% \begin{figure}[h!]
% 	\includegraphics[width=0.99\linewidth]{figures/INSPComparison}
%         \vspace{-.3cm}
% 	\caption{
% 	 A qualitative comparison between INSP \cite{xu2022signal}, BACON \cite{lindell2022bacon} and our approach using a Gaussian kernel.
%         INSP approach does hardly perform any filtering, when the kernel is larger, such as our approach supports, but suffers from noise.
%         BACON fairs better, but shows ringing at high-contrast edges such as the house against the sky.
%         \TODO{add PNF results}
% 	}
% 	\label{fig:insp_comparison}
% \end{figure}

\begin{table}[]
    \centering
    \caption{Image quality comparison for Gaussian kernels.}
    %\vspace{-0.4cm}
    \renewcommand{\tabcolsep}{0.095cm}
    \label{tab:comparison}
\begin{tabular}{lrrrrrrrrr}
& \multicolumn{3}{c}{$\sigma=0.04$} & \multicolumn{3}{c}{$\sigma=0.05$} & \multicolumn{3}{c}{$\sigma=0.07$} \\
\cmidrule(lr){2-4}
\cmidrule(lr){5-7}
\cmidrule(lr){8-10}
& \footnotesize{PSNR} & \footnotesize{LPIPS} & \footnotesize{SSIM} & \footnotesize{PSNR} & \footnotesize{LPIPS} & \footnotesize{SSIM} & \footnotesize{PSNR} & \footnotesize{LPIPS} & \footnotesize{SSIM} \\
\toprule
INSP &	27.2 &	0.305 &	0.774 &	25.8 &	0.398 &	0.713 &	23.9 &	0.535 &	0.622 \\
BACON &	\winner{35.2} &	0.079 &	\winner{0.946} &	33.6 &	0.091 &	0.935 &	30.6 &	0.139 &	0.904 \\
PNF &	34.5 &	\winner{0.072} &	0.938 & \winner{33.9} &	0.086 &	0.935 &	32.0 &	0.132 &	0.922 \\
\textbf{Ours} &	28.6 &	0.076 &	0.934 &	31.0 & \winner{0.042} &	\winner{0.963} &	\winner{34.1} &	\winner{0.032} & \winner{0.98} \\
\bottomrule
\end{tabular}
\end{table}

\begin{figure}[h!]
	\includegraphics[width=0.99\linewidth]{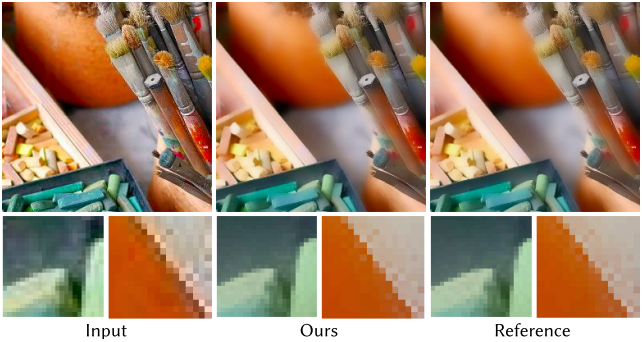}
        %\vspace{-.3cm}
	\caption{
        Non-linear (bilateral) filtering of an input image (left) by our method (middle) and the reference (right).}
 %\vspace{-.3cm}
	\label{fig:bilateral}
\end{figure}

\paragraph{Non-linear filtering}
Our approach can also be used for non-linear filtering, such as bilateral filtering \cite{tomasi1998bilateral}.
We implement this by optimizing for the repeated integral field of a bilateral grid \cite{chen2007real}, which augments a 2D image with an additional signal dimension, reducing filtering to a linear operation in this extended space.
We see in Fig.~\ref{fig:bilateral} that our approach is able to faithfully produce edge-aware filtering results.

\begin{figure}[h!]
	\includegraphics[width=0.99\linewidth]{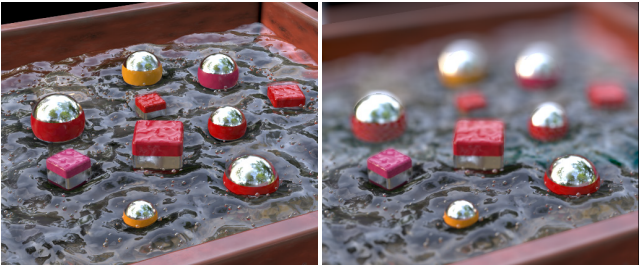}
    % \vspace{-0.2cm}
	\caption{Depth-of-field applied to a synthetic image with a depth map.}
	\label{fig:gbuffer}
\end{figure}

\paragraph{Spatially-varying filtering}
is demonstrated in Fig.~\ref{fig:teaser} and Fig.~\ref{fig:gbuffer}.
In these cases, the strength of a circular blur filter is determined by a spatially-varying auxiliary signal in the form of a depth buffer.

% \begin{figure}[h!]
% 	\includegraphics[width=0.99\linewidth]{figures/Bilateral}
%         \vspace{-.3cm}
% 	\caption{
%         Non-linear (bilateral) filtering of an input image (left) by our method (middle) and the reference (right).
%         \TODO{update images}
% 	}
%  \vspace{-.3cm}
% 	\label{fig:bilateral}
% \end{figure}

% \begin{figure}[h!]
% 	\includegraphics[width=0.99\linewidth]{figures/MotionBlur}
%         \vspace{-.3cm}
% 	\caption{
% 	 We apply our approach to a 3D space-time HDR field (video, seen left) to filter along the time axis with a Gaussian kernel.
%   The resulting motion blur (middle) compares favorably to the reference.
% 	}
%  \vspace{-.3cm}
% 	\label{fig:motion_blur}
% \end{figure}

%==========================================

\subsection{Videos}
\begin{figure}[h!]
	\includegraphics[width=0.99\linewidth]{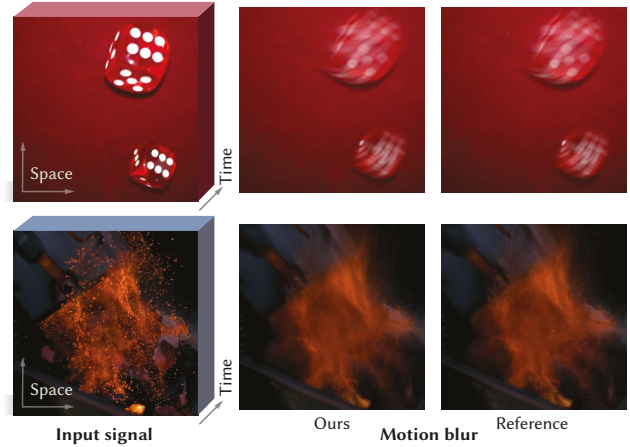}
        %\vspace{-.3cm}
	\caption{
	 We apply our approach to a 3D space-time HDR field (video, seen left) to filter along the time axis with a tent kernel.
  The resulting motion blur (middle) compares favorably to the reference.
	}
 %\vspace{-.3cm}
	\label{fig:motion_blur}
\end{figure}

Videos are fields $\field \in \mathds{R}^3 \rightarrow \mathds{R}^3$, mapping 2D location and time to RGB color.
In Fig.~\ref{fig:motion_blur}, we apply smoothing with a tent filter along the time dimension to create appealing non-linear motion blur.
We refer to our supplemental material for a more extensive evaluation.

% \begin{figure}[h!]
% 	\includegraphics[width=0.99\linewidth]{figures/Gbuffer}
%         \vspace{-.3cm}
% 	\caption{
% 	 Depth-of-field applied to a synthetic image with a depth map.
% 	}
%  \vspace{-.3cm}
% 	\label{fig:gbuffer}
% \end{figure}

%==========================================

\subsection{3D Geometry}

\begin{figure*}[htb]
        \includegraphics[width=0.99\linewidth]{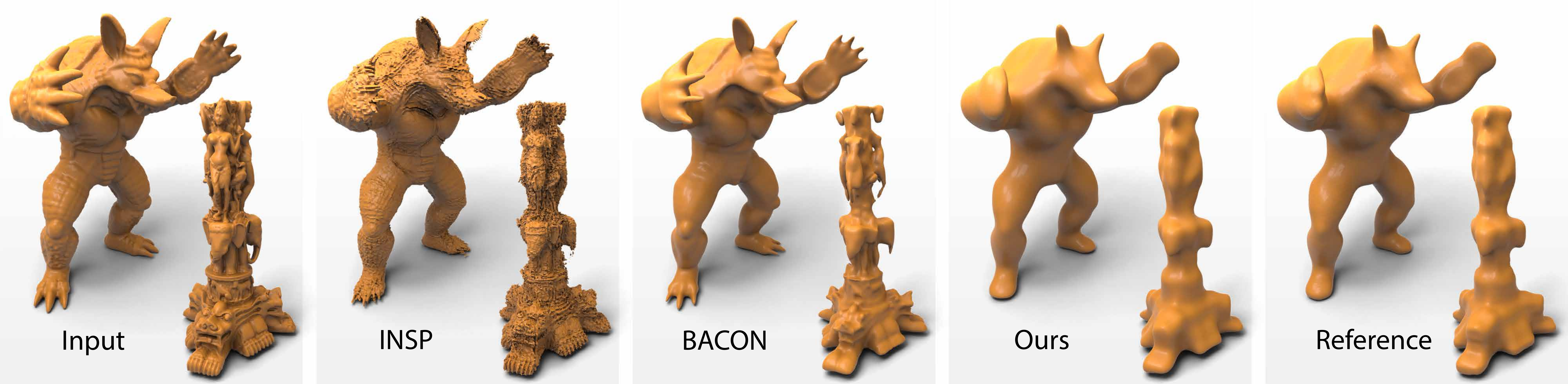}
    % \vspace{-.2cm}
	\caption{
	  Two geometric shapes represented by an SDF (left) are filtered with a box kernel ($\sigma = 0.05$).
   While INSP, in the second column, suffers from noise, and BACON, in the third column, cannot reproduce larger-scale filtering, our result is close to the MC reference.
    	}
 %\vspace{-.3cm}
	\label{fig:geometry}
\end{figure*}

Surfaces can be modeled using signed distance functions (SDFs)
$\field \in \mathds{R}^3 \rightarrow \mathds{R}$.
We apply 3D box filters with different side lengths $\sigma$ to an SDF, resulting in a progressively smoothed surface.
%Low-pass filtering of an SDF naturally results in morphological simplifications.
We again compare our approach to INSP and BACON (no code is available to apply PNF to the 3D case) in Fig.~\ref{fig:geometry} and Table \ref{tab:comparison_geometry}, where numerical evaluations are averaged across the three 3D objects studied in INSP.
As metrics, we consider the mean squared error (MSE) of the SDF, the intersection over union (IoU), as well as the Chamfer distance of the reconstructed surface.
We observe that we outperform INSP and BACON across all kernel sizes and metrics.
More qualitative results can be found in supplemental.
Additionally, Fig.~\ref{fig:spatially_varying_3D} demonstrates a spatially-varying SDF filtering result.

\begin{figure}[h!]
	\includegraphics[width=0.99\linewidth]{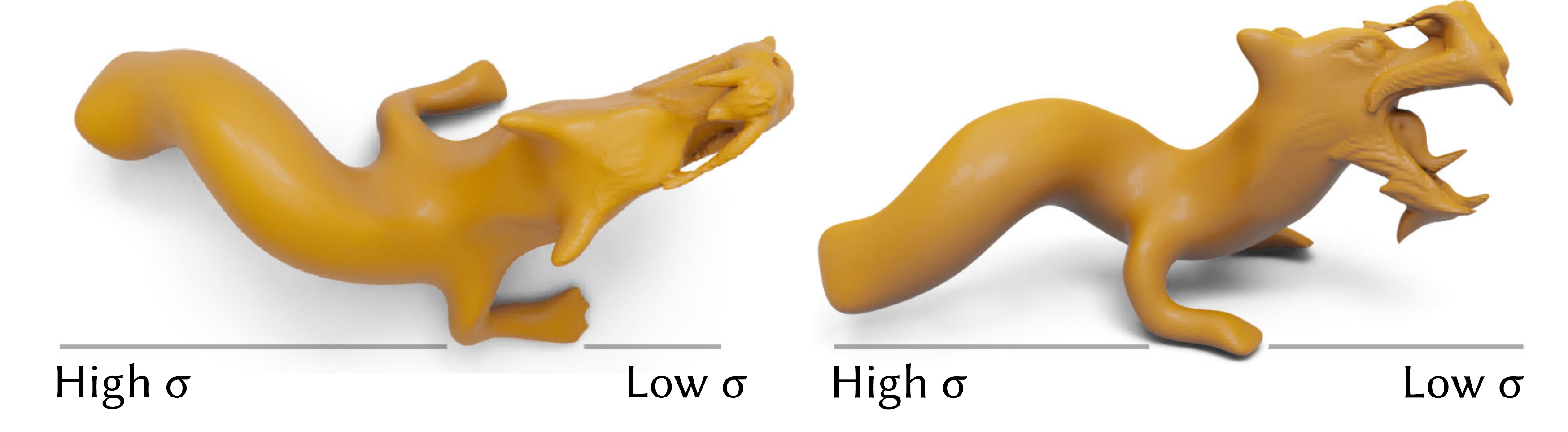}
        %\vspace{-.3cm}
	\caption{
	  Spatially-varying blur from two views computed using our method.
    	}
 %\vspace{-.3cm}
	\label{fig:spatially_varying_3D}
\end{figure}

% \begin{figure}[h!]
% 	\includegraphics[width=0.99\linewidth]{figures/geometry}
%         \vspace{-.2cm}
% 	\caption{
% 	  Two geometric shapes represented by an SDF (left) are filtered with a box kernel $\sigma = 0.05$).
%    While INSP, in the second column, suffers from noise, our result is close to the MC reference.
%    \TODO{update images, also put BACON results}
%     	}
%  \vspace{-.3cm}
% 	\label{fig:geometry}
% \end{figure}

% \begin{figure}[h!]
% 	\includegraphics[width=0.99\linewidth]{figures/SpatiallyVarying3D}
%         \vspace{-.2cm}
% 	\caption{
% 	  Spatially-varying blur across a dragon from two views computed using our method.
%    \TODO{add axis}
%     	}
%  \vspace{-.3cm}
% 	\label{fig:spatially_varying_3D}
% \end{figure}

\begin{table}[]
    \centering
    \caption{Quality comparisons of filtered SDFs using 3D box kernels.}
    \vspace{-0.4cm}
    \renewcommand{\tabcolsep}{0.18cm}
    \label{tab:comparison_geometry}
\begin{tabular}{
    rrrrrrr}
& 
\multicolumn{3}{c}{$\sigma=0.05$}&
\multicolumn{3}{c}{$\sigma=0.15$} \\
\cmidrule(lr){2-4}
\cmidrule(lr){5-7}
&
\multicolumn1c{\footnotesize{MSE}}& 
\multicolumn1c{\footnotesize{Cham.}}&
\multicolumn1c{\footnotesize{IoU}}& 
\multicolumn1c{\footnotesize{MSE}}&
\multicolumn1c{\footnotesize{Cham.}}&
\multicolumn1c{\footnotesize{IoU}}
\\
\toprule
INSP & 292.50000 & 171.77 & 0.90 & 281.30000 & 388.26 & 0.73 \\
BACON & 0.40145 & 360.90 & 0.82 & 1.85791 & 932.84 & 0.61 \\
Ours & 
\winner{0.00109} & \winner{20.97} & \winner{0.99}& \winner{0.00026} & \winner{13.24} & \winner{0.99} \\
\bottomrule
\end{tabular}
\end{table}

%==========================================

\subsection{Animation}

\begin{figure}[h!]
	\includegraphics[width=0.99\linewidth]{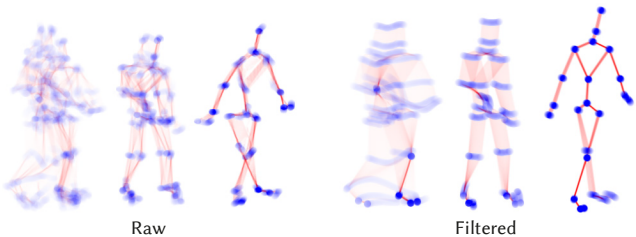}
        % \vspace{-.3cm}
	\caption{
	 A noisy motion-capture sequence (left) is filtered using our approach to yield smooth motion trajectories (right).
	}
	\label{fig:animation}
\end{figure}

We consider the task of filtering a neural field representation of a motion-capture sequence, which contains a significant amount of noise.
Our test sequence consists of 23 3D joint position paths over time, resulting in a field
$\field \in \mathds{R} \rightarrow \mathds{R}^{69}$.
In Fig.~\ref{fig:animation} and the supplemental, we show the result of applying a Gaussian filter to the noisy animation data, resulting in smooth motion trajectories.

% \begin{figure}[h!]
% 	\includegraphics[width=0.99\linewidth]{figures/animation}
%         \vspace{-.3cm}
% 	\caption{
% 	 A noisy motion-capture sequence (left) is filtered using our approach to yield smooth motion trajectories (right).
% 	}
%  \vspace{-.3cm}
% 	\label{fig:animation}
% \end{figure}

%==========================================

\subsection{Audio}

Finally, we apply our framework to the task of filtering an audio signal $\field \in \mathds{R} \rightarrow \mathds{R}$, available for listening in the supplemental.

%==========================================

\section{Analysis}

Here, we analyze further individual aspects of our approach.

\begin{figure*}
	\includegraphics[width=0.99\linewidth]{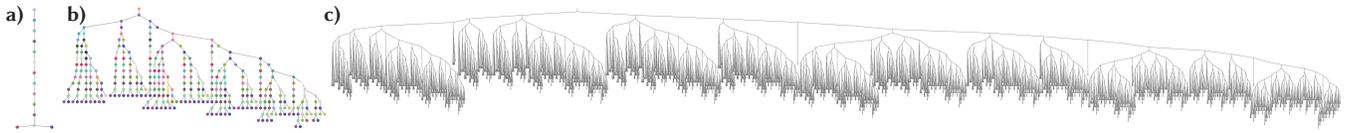}
    % \vspace{-0.3cm}
	\caption{
	\revision{Comparison of different graph sizes.
    \emph{a)} Our method retains a small graph independent of integration order.
    \emph{b)} The AutoInt graph after two differentiations (first Int. Op. in Tab.~\ref{tab:integral_numerics}).
    \emph{c)} The AutoInt graph after four differentiations (second Int. Op. in Tab.~\ref{tab:integral_numerics}). Same colors represent same operations.}
	}
	\label{fig:autoint}
\end{figure*}

\paragraph{Repeated Integral Fields}
\label{sec:repeated_integral_fields}
We seek to gain more insights into our learned integral fields.
To this end, we consider the 2D image case (Sec.~\ref{sec:images}) both for single and double integrals per dimension, as required for convolutions with piecewise constant and linear kernels, respectively.
In Tab.~\ref{tab:integral_numerics} we compare our antiderivatives against AutoInt \cite{lindell2021autoint}.
\revision{As the original implementation does not support integration with respect to more than one variable, we re-implemented this baseline using the \texttt{functorch} library for repeated differentiation.}
We compute the mean squared error (MSE) between the original signal and the obtained integral fields after repeated automatic differentiation, averaged over three images.
We do not compare integrals directly, as their absolute values are dominated by higher-order constants of integration.
Further, we measure the time required to train the fields, as well as the size of the network during training in terms of the number of nodes.
\revision{As there is no straightforward procedure to count the number of nodes of the repeated-derivative graphs, we use the official AutoInt implementation for this particular calculation and differentiate two and four times with respect to \emph{one} input variable, to get a good approximation of the two cases studied.
Corresponding graphs are visualized in Fig.~\ref{fig:autoint}.}

We see that our approach produces higher-quality antiderivatives than AutoInt while taking significantly less time to train, in particular for higher-order integrals. Further, our network size during training is independent of the integration order, while the computational graphs of AutoInt grow quickly due to the symbolic differentiation required.

\revision{We are further interested in how the quality of the learned antiderivative affects convolution quality. 
For this analysis, we consider antiderivative MSE as above (lower is better), and convolution quality in terms of PSNR (higher is better). We find the two measures highly correlated: Pearson’s R = -0.98.}

In appendix C of the supplemental document, we study the accuracy of our optimized kernels.

\begin{table}[]
    \centering
    \caption{Integral field evaluation.}
   % \vspace{-0.3cm}
    \label{tab:integral_numerics}
\begin{tabular}{clrrr}

\toprule
Int. Op. & Method & MSE\,\footnotesize{($\times 10^{-3}$)} & Time & Graph Size \\
\midrule

\multirow{2}{*}{$\int \!\! \int$} 
& AutoInt & 6.83 & 1.3h & 339 \\
& \textbf{Ours} & \winner{6.48} & \winner{1.1h} & \winner{11} \\

\midrule

\multirow{2}{*}{$\int^2 \!\! \int^2$} 
& AutoInt & 7.71 & 14.7h & 15,407 \\
& \textbf{Ours} & \winner{7.28} & \winner{1.2h} & \winner{11} \\

\bottomrule
\end{tabular}
\end{table}

\begin{figure}[h!]
    \includegraphics[width=1\linewidth]{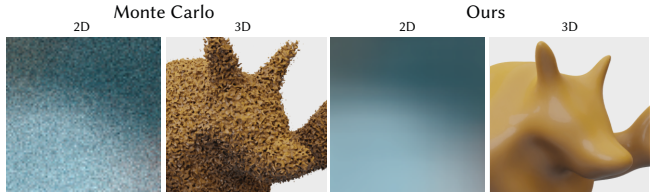}
    \caption{Ours vs.\ an equal-effort Monte Carlo estimate of a  convolution.}
    
    \label{fig:mc_vs_ours} 
\end{figure}

\paragraph{Comparison to equal-effort MC}
While we use \emph{converged} Monte Carlo estimates of convolutions (Eq.~\ref{eq:mc_conv}) as references in our experiments, we are also interested in the quality of such an estimate when reducing the number of samples to the number of Dirac deltas we use in our method, providing an equal-effort comparison.
In Fig.~\ref{fig:mc_vs_ours} we show an illustration of this analysis.
We observe that MC suffers from extensive noise, while our solution is smooth.

%==========================================

\section{Discussion and Conclusion}

We have presented a novel approach to perform general, spatially-varying convolutions in continuous signals.
Capitalizing on the fact that piecewise polynomial kernels become sparse after repeated differentiation, we only require a small number of integral-network evaluations to perform large-scale continuous convolutions.

Since our work is one of the first steps in this direction, there is ample opportunity for future work. 
Currently, one of our biggest limitations is that we need access to the entire signal to train our repeated integral field.
This prevents the treatment of signals that are only partially observed through a differentiable forward map, e.g., as prominently is the case for neural radiance fields \cite{mildenhall2020nerf}.

Our integral fields are trained using generalized finite differences, which we found to become unstable for small kernels (Sec.~\ref{sec:implementation}).
This naturally imposes an upper frequency limit on the learned antiderivatives.
Fortunately, this becomes noticeable only for small kernels, in which case we resort to Monte Carlo sampling of the convolution, which is efficient in this condition.

Our assumption is that kernels to be used for spatially-varying filtering are (anisotropically) scaled versions of a reference kernel, which, arguably, covers a broad range of applications.
We do not have a scalable solution for situations that violate this assumption -- except for special problems such as bilateral filtering. 
One solution could be blending between the Dirac deltas of a set of pre-computed reference kernels.
Further, kernel transformations are limited to axis-aligned operations.
We envision that re-parameterizations leveraging the continuous nature of the representation might be able to lift this restriction. 

\revision{In this work, we do not claim superiority over operating on a grid, if enough space is available to represent the signal that way. But if one needs to use a neural field, e.g. for extreme compression, continuous queries, or other advantages (Sec.~\ref{sec:introduction}), an approach such as ours is needed.}
We hope to inspire future work on signal processing in continuous neural representations to help them reach their full potential.

%\begin{acks}
%xxx
%\end{acks}

% ----------------------------------------------------------------

\bibliographystyle{ACM-Reference-Format}
\bibliography{paper}

% ----------------------------------------------------------------

\clearpage
\appendix
\section{Convolution Methods}
\label{sec:convolution_methods}

In Tab.~\ref{tab:convolution_methods} we compare different solutions to the convolution problem.

\begin{table}
    \centering
     \caption{Different convolution methods. $n$ is the size of the filter, $m$ the size of the signal (samples or weights to represent it),  and $d$ the signal dimension.}
     %\vspace{-0.3cm}
    \begin{tabular}{llccc}
         &
         \multicolumn1c{Time}&
         \multicolumn1c{Spat. vary}&
         \multicolumn1c{Noisy}&
         \multicolumn1c{Cont.}\\
         \toprule
         Classic&
         $O(m\times n^d)$&
         \cmark&
         \xmark&
         \xmark\\
         Fourier&
         $O(m\times \log(m)\times d)$&
         \xmark&
         \xmark&
         \xmark\\
         Monte Carlo&
         $O(m \times n)$&
         \cmark&
         \cmark&
         \cmark\\
         SAT&
         $O(m\times d)$&
         \cmark&
         \xmark&
         \xmark\\
         Mip-NeRF&
         $O(m)$&
         \cmark&
         \xmark&
         \cmark\\
         INSP&
         $O(m)$&
         \xmark&
         \xmark&
         \cmark\\
         \textbf{Ours}&
         $O(m)$&
         \cmark&
         \xmark&
         \cmark\\
         \bottomrule
    \end{tabular}
    \label{tab:convolution_methods}
\end{table}

\section{Integral Field Model Details}
\label{sec:ModelDetails}

In Tab.~\ref{tab:models} we give details of network architectures for the repeated integral fields we use per application.
In all cases we use a multi-layer perceptron (MLP).
Similar to \citet{lindell2021autoint}, we observed best results with Swish \cite{ramachandran2017searching} activation functions.
We report the number of hidden layers, the number of features per layer, and the resulting number of trainable parameters.

\begin{table}[]
    \centering
    \caption{Architecture details of our integral fields.}
    \label{tab:models}
     \begin{threeparttable}
    \begin{tabular}{lrrr}
        \toprule
        Application & \#Layers & \#Features & \#Trainable Params. \\
        \midrule
        Images\tnote{1} & 5 & 256 & 270,851 \\
        Images\tnote{2} & 5 & 512 & 1,065,987 \\
        Videos & 9 & 256 & 534,019 \\
        Geometry & 5 & 256 & 270,851 \\
        Audio & 5 & 256 & 270,851 \\
        Animation & 5 & 256 & 270,851 \\
        \bottomrule
    \end{tabular}
    \begin{tablenotes}\footnotesize
        \item[1] Low-resolution (256x256) images used for large-scale comparisons.
        \item[2] High-resolution (3000x3000) images used for displaying results.
    \end{tablenotes}
    \end{threeparttable}
\end{table}

\section{Kernels}
\label{sec:kernel_accuracy}

In Tab.~\ref{tab:kernel_data} we provide details on our optimized kernels.
Here, we consider a 1D Gaussian kernel, represented by different numbers \diraccount of Diracs, using different orders of differentiation \intcount.
We give the reconstruction error in terms of the mean squared error (MSE) and the time (in seconds) our unoptimized implementation takes to obtain a converged result.

\begin{table}[]
    \centering
    \caption{Accuracy of kernels and time to optimize them.}
    \label{tab:kernel_data}
    \renewcommand{\tabcolsep}{0.09cm}
    \scalebox{0.95}{
\begin{tabular}{lcccccccc}
\diraccount & \multicolumn{2}{c}{3} & \multicolumn{2}{c}{7} & \multicolumn{2}{c}{13} & \multicolumn{2}{c}{24} \\
\cmidrule(lr){2-3} 
\cmidrule(lr){4-5}
\cmidrule(lr){6-7}
\cmidrule(lr){8-9}
\intcount & 1 & 2 & 1 & 2 & 1 & 2 & 1 & 2 \\
\toprule
MSE & 1.6e-1 & 1.5e-2 & 1.5e-2 & 7.9e-4 & 4.0e-3 & 7.6e-5 & 1.1e-3 & 2.3e-5 \\
Time\,(s) & 3 &	25 & 4 & 60 & 6 & 75 & 9 & 132 \\
\bottomrule
\end{tabular}
}
\end{table}

\end{document}